\documentclass[conference, letterpaper]{IEEEtran}
\usepackage{cite}
\usepackage{amsmath,amssymb,amsfonts}
\usepackage{algorithmic}
\usepackage{graphicx}
\usepackage{textcomp}
\def\BibTeX{{\rm B\kern-.05em{\sc i\kern-.025em b}\kern-.08em
    T\kern-.1667em\lower.7ex\hbox{E}\kern-.125emX}}

\usepackage{float}

\usepackage{changes}
\definechangesauthor[name=alban, color=blue]{al}

\begin{document}

\title{Learning Representations of Spatial Displacement through Sensorimotor Prediction }

 \author{
 Michael Garcia Ortiz$^1$, 
 Alban Laflaqui\`ere$^1$, 
 \\ 
 $^1$ AI Lab, SoftBank Robotics Europe\\
 %
 mgarciaortiz@softbankrobotics.com,
 alaflaquiere@softbankrobotics.com,
 }

\maketitle

\begin{abstract}
  Robots act in their environment through sequences of continuous motor commands. Because of the dimensionality of the motor space, as well as the infinite possible combinations of successive motor commands, agents need compact representations that capture the structure of the resulting displacements. In the case of an autonomous agent with no a priori knowledge about its sensorimotor apparatus, this compression has to be learned. 
We propose to use Recurrent Neural Networks to encode motor sequences into a compact representation, which is used to predict the consequence of motor sequences in term of sensory changes. We show that sensory prediction can successfully guide the compression of motor sequences into representations that are organized topologically in term of spatial displacement.

  
\end{abstract}

\section{Introduction}

 The goal of Developmental Robotics \cite{Cangelosi:2014:DRB:2628000} is to endow agents with the capability to learn how to act and interact in their environment. With minimal a priori knowledge and bias provided by the designer, the agent has to progressively build knowledge and skills to interact with its environment. It is a particularly desirable property for the case of open-ended learning, when the agent does not know in advance what situations it might encounter. 
This knowledge can take the form of a model which condenses past experiences and enables prediction of the outcome of future actions. Improving the capability of the agent to predict the consequences of its actions means refining the model of the world.
This vision is in line with the modern Predictive Coding Theory of Cognition which describes brains as predictive machines guided by prediction error minimization (see, e.g., \cite{Friston1211}).
 For any non-trivial agent, a major difficulty in the autonomous learning of a predictive model is the dimensionality of the sensorimotor space. It is generally higher than the effective "task space" in which the agent can actually interact with the world.
We investigate if this high dimensional space can be encoded into a low dimensional representation through sensorimotor prediction.

We illustrate the need for compact motor sequences representation with a simulated navigation scenario. When an agent moves in space through a sequence of motor commands, it changes its position and orientation in the world. In order to perform accurate sensorimotor prediction, the agent needs to learn this space of displacement. 
The number of motor command sequences that a mobile robot can generate to move around in its environment is significantly greater than the number of displacements it can effectively undergo. An infinite number of trajectories in space can lead to a similar change of position and orientation. 
Without a priori knowledge, the agent has only access to the high-dimensional space of motor sequences but not to its low-dimensional equivalent: the space of displacements in the environment. 
This internal representation of motor sequences should capture the structure of a navigation task. In particular, the different motor sequences that correspond to identical displacements should be encoded by the same internal representation.

 We propose an unsupervised (or self-supervised) approach based on sensorimotor prediction to build an internal representation of displacement. 
 In order to encode a sequence of motor commands, we propose to use a Recurrent Neural Network (RNN)  to encode this sequence into a compact space, which is used to predict the sensory values after the displacement.  
 In such a framework, the structure of sensory experiences acts as a proxy for the structure of the task and guides the building of the motor representations. Regularities experienced in the sensory flow are thus used to compress motor information.




\section{Related work}

In \cite{LAFLAQUIERE201549}, the authors studied the space of displacement of an agent consisting of a multi-joint arm equipped with a retina-like sensor. It experiences a variety of poses corresponding to multiple motor configurations, in front of multiple light sources. They showed that the intrinsic space of configurations of the sensor could be compacted into a topological low dimensional representation where redundant configurations of the arm would collapse to a single state.
We took inspiration from the general idea of using sensory regularities to build motor topologies, and extended it to learning representations of motor sequences leading to displacements in an environment, where the agent not only modifies its point of view but also its position with regard to the external world.

Learning forward models \cite{Wolpert19961265} is a classical way to obtain sensorimotor predictive models for an agent. Recently, numerous works proposed to learn forward models in order to acquire predictive ability and improve the capacity of agent to build useful sensor state representations.
In \cite{DBLP:journals/corr/MirowskiPVSBBDG16}, the authors propose to use prediction and loop closure detection as auxiliary tasks to better solve a navigation problem in a Reinforcement Learning framework. They showed that enforcing the capacity to predict future states helped developing useful representations and policies for navigation.
Similarly, in \cite{DBLP:journals/corr/AgrawalNAML16}, the authors successfully modeled the dynamics of a real robotic arm by learning an inverse and a forward model through exploration. The robot could learn useful and informative visual representations, improving the ability of the robot to solve manipulation tasks.
In \cite{DBLP:journals/corr/abs-1709-05185}, the authors extended the approach from \cite{jonschkowski2015learning}, and showed that state representations for robotics benefits from imposing particular properties on the representation space (for example, continuity and proportionality).
These approaches are mainly concerned with the learning of sensor state representations (see \cite{DBLP:journals/corr/abs-1802-04181} for a complete overview), and assume very simple motor commands.

All these contributions indicate that it is possible to build spatial representations through prediction. If most works are concerned with sensor state representation, recent contributions focus on learning representations of spatial displacement.
In \cite{j.2018emergence}, the authors propose an experiment where an agent moves by rotations and forward movements, and perform path integration. It continuously predicts its Cartesian coordinates using a RNN, and the authors showed that the state of the RNN contained information about spatial displacement. The agent has access to explicit Cartesian coordinates to perform path integration, which is not available for animals or autonomous mobile agents.
Similarly, in \cite{hippopred}, the compression of a predictive model lead to the emergence of representations akin to grid cells, in the case of an agent moving along a triangular lattice.
The discrete motor space doesn't allow to generalize and draw conclusions on the ability of the agent to learn representations for continuous displacements.

To our knowledge, this paper is the first account of learning representations of spatial displacement based on sequences of motor commands, in a rich sensorimotor environment.

\section{Learning Representations of motor sequences}
 
\begin{figure}[h!]
\centering
\includegraphics[width=0.49\textwidth]{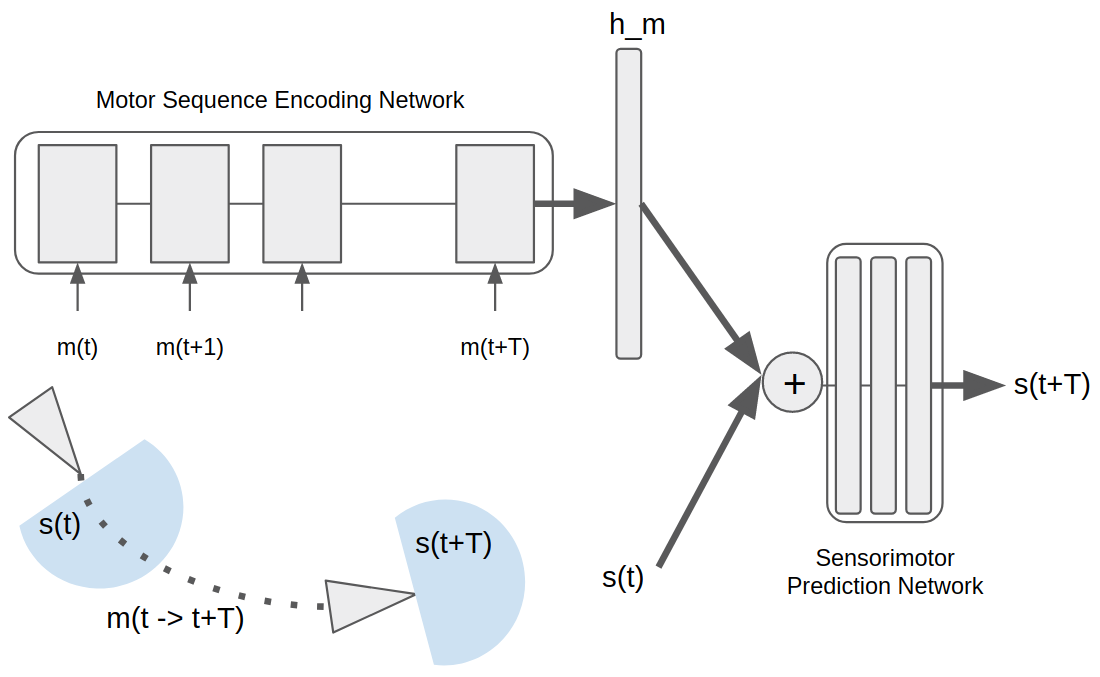}
\caption{Architecture: the motor sequence encoding network encodes motor sequences into a representation $\mathbf{h^m}$, which is used for sensorimotor prediction }
\label{fig:archi}
\end{figure}

We assume that an agent moves in an environment by performing sequences of individual motor commands $\mathbf{m} = \{ m^0, m^1, ... , m^N \} $ on N different actuators. It captures information about its environment through an ensemble of K sensors $\mathbf{s} = \{ s^0, s^1, ... , s^K \} $.
At each timestep $t$, the agent performs a new motor command $ \mathbf{m}_t $ and receives new sensor values $ \mathbf{s}_t $.
The goal of the sensorimotor prediction is to learn a function that predicts future sensory values $\mathbf{s}_{t+T}$ based on the current sensory values $\mathbf{s}_t$ and a sequence of motor commands $[ \mathbf{m}_{t+1}, \mathbf{m}_{t+2}, ... , \mathbf{m}_{t+T} ]$.
We propose to learn this function using Neural Networks, which are a generic function approximator.
Our system, illustrated in Fig.\ref{fig:archi}, is composed of a \textit{motor sequence encoding network} and a \textit{sensorimotor prediction network}. The first one  compresses a sequence of motor commands into a compact representation $\mathbf{h^m}([m_{t},\dots,m_{t+\tau}])$ (abbreviated $\mathbf{h^m}$). $\mathbf{h^m}$ is concatenated with $\mathbf{s}_t$  and used by the \textit{sensorimotor prediction network} to predict $\mathbf{s}_{t+T}$.

\subsubsection{Motor sequence encoding network}

The motor sequence encoding network takes as  input a sequence of motor commands 
$ [ \mathbf{m}_{t+1}, ..., \mathbf{m}_{t+2},\mathbf{m}_{t+T} ] $, where t is the current timestep, and T is the number of motor commands in the sequence, or the prediction horizon. The sequence is fed to a Long Short-Term Memory (LSTM) \cite{hochreiter1997long}, an instance of RNNs, wich at each timestep updates its internal state with the previous state and the new motor command (the state of the LSTM is initialized to 0). The final state is projected on the motor representation layer $\mathbf{h^m}$, which is a a sigmoid layer fully connected to the last state of the LSTM. The vector of activation $\mathbf{h^m}$ constitutes the motor sequence representation that is used for sensorimotor prediction over a sequence.

\subsubsection{Sensorimotor prediction network}

$\mathbf{h^m}$ is concatenated with the current sensor value $\mathbf{s}_{t}$ and fed to 3 layers of fully connected Rectifier Linear Units (ReLu), which role is to predict the future sensor value $\mathbf{s}_{t+T}$. It outputs a prediction $\mathbf{s}^*_{t}$, which is compared to the actual ground truth to train the network.
The system is trained end-to-end, by minimizing the mean squared error of prediction:
\begin{equation}
L =  \mathbb{E} [ | \mathbf{s}^*_{t+T} - \mathbf{s}_{t+T}|^2 ] 
\end{equation}

\section{Test Environments}
\subsection{Lattice world as a baseline}

As a baseline, we use a simple discrete environment, where an agent lives in a 6x6 lattice. It changes its position, corresponding to a node of the lattice, by performing at each timestep a random rotation $\theta_{body} \in \{-\pi/2, 0, \pi/2\}$,  followed by a random forward movement $\delta_{forward} \in \{0, 1\}$. During exploration, any movement outside of the lattice is discarded.
The agent receives as $\mathbf{s}_t$ the index of the node which it is visiting and its absolute orientation. It predicts the future index of the node where it will end up after a sequence of $T$ movements. The index of the grid, the absolute orientation and the motor values are all encoded as one-hot vectors. The agent therefore has no explicit access to spatial information. 

\subsection{ Continuous world with a forward Agent} \label{sec:forward_agent}

\begin{figure}[h!]
\centering
\includegraphics[width=0.25\textwidth]{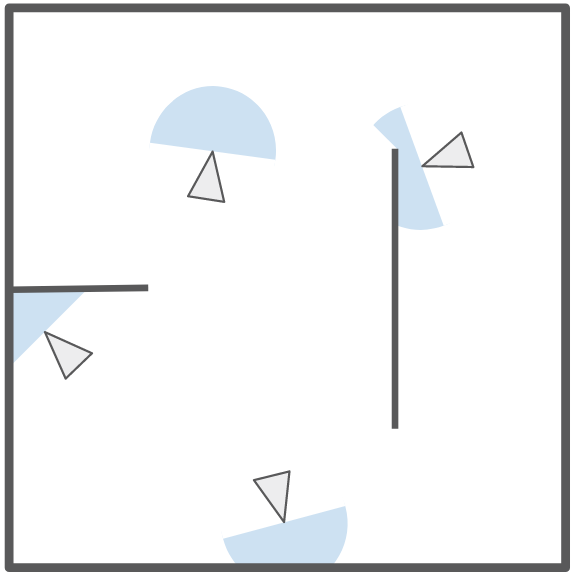}
\caption{Illustration of the continuous environment used for our simulations}
\label{fig:env}
\end{figure}

We propose an agent which lives in a continuous world of size 40x40 composed of external and internal walls, illustrated in Fig.\ref{fig:env}.
The Forward Agent performs at each timestep, a random rotations  $\theta_{body}$ sampled uniformly in  $[ -\theta^{max}_{body}, \theta^{max}_{body} ]$, followed by a random forward movement  $\delta_{forward} $ sampled uniformly in  $ [0, \delta^{max}_{forward} ] $.
During the exploration, any movement that would lead the agent through a wall is discarded.
The agent is equipped with 9 distance sensors of range 10, each providing a real value corresponding to the distance to an obstacle. They are equally spaced in its field of view of angle $\pi$ centered on the longitudinal axis of the agent.

\subsection{ Continuous world with a holonomic agent}\label{sec:holo_agent}

In order to study the learning of motor sequences representation in cases where the motor space is redundant, we propose a holonomic Agent, moving in the same continuous environment as in Sec.\ref{sec:forward_agent}. The agent can translate on its longitudinal and lateral axis, and can rotate its body. Additionally, it is equipped with a head on which the sensors are attached. The field of view of the agent is centered on the orientation of the head.
It can change the orientation of its head by changing the angle of its head relative to its body. 
At each timestep, the agent performs successively:
\begin{itemize}
\item a random rotation of its body $\theta_{body}$ sampled uniformly   in $[ -\theta^{max}_{body}, \theta^{max}_{body} ]$ ,
\item a random longitudinal translation of its body $\delta_{long}$  sampled uniformly   in $ [ -\delta_{long}^{max}, \delta_{long}^{max} ]$ ,
\item a random lateral translation of its body $\delta_{lat}$  sampled uniformly   in $ [ -\delta_{lat}^{max}, \delta_{lat}^{max} ]$ ,
\item a random change of its head angle relative to its body  $\theta_{head}$ sampled uniformly  in  $[ -\theta^{max}_{head}, \theta^{max}_{head} ]$ .
\end{itemize}
This Holonomic Agent allows to evaluate our approach on motor spaces which are redundant, and of different nature (setting the relative head orientation in contrast with changing the absolute body orientation by a certain angle) compared to the Forward agent.

\begin{figure}[h!]
\centering
\includegraphics[width=0.3\textwidth]{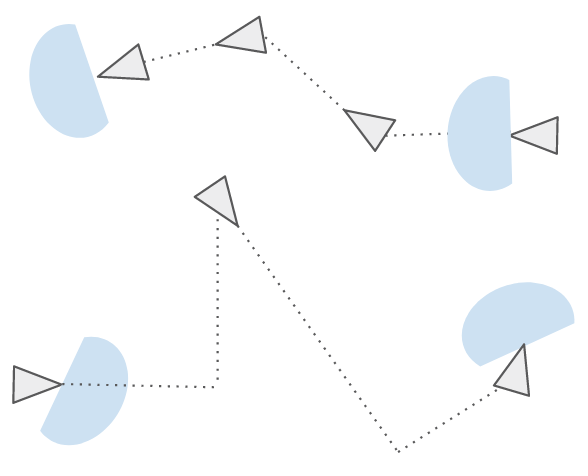}
\caption{Example trajectories for Forward (top) and Holonomic (bottom) Agent }
\label{fig:traj}
\end{figure}


\section{Experiments and Results}

For each experiment presented in this section, we learn the sensorimotor prediction, and subsequently the motor sequence encoding, by sampling random motor sequences from random initial position, body orientation and head orientation (when applicable) in the environment. We collect training samples $ \{[ \mathbf{s}_t, [ \mathbf{m}_{t+1}, ..., \mathbf{m}_{t+2},\mathbf{m}_{t+T} ], \mathbf{s}_{t+T} \}$ and train our system to perform future sensory prediction.
We use the same network parameters for all of our experiments: the LSTM has a state size of 100, and the layers for the prediction have a size of 200. The representation layer is a sigmoid layer of size 50. The network is trained end-to-end using ADAM optimizer. 

In order to visualize the representation space $\mathbf{h^m}$ , once the system is trained, we sample $10e5$  random trajectories, and project their representations $\mathbf{h^m}$  using  the first components of their Principal Component Analysis (PCA). Knowing where the agent ended after a sequence of motor command, we can assign a color to this point corresponding to its relative longitudinal displacement, relative lateral displacement, or relative orientation change. 
For all visualization, we use the jet color code normalized on the projected data, where red corresponds to a maximal value, and blue to a  minimal value.
We plot all the points with assigned color codes and observe the structure of the learned representation space $\mathbf{h^m}$ .

\begin{figure*}[t!]
\noindent

\begin{minipage}[t]{0.24\textwidth}
\centering
\includegraphics[width=\textwidth]{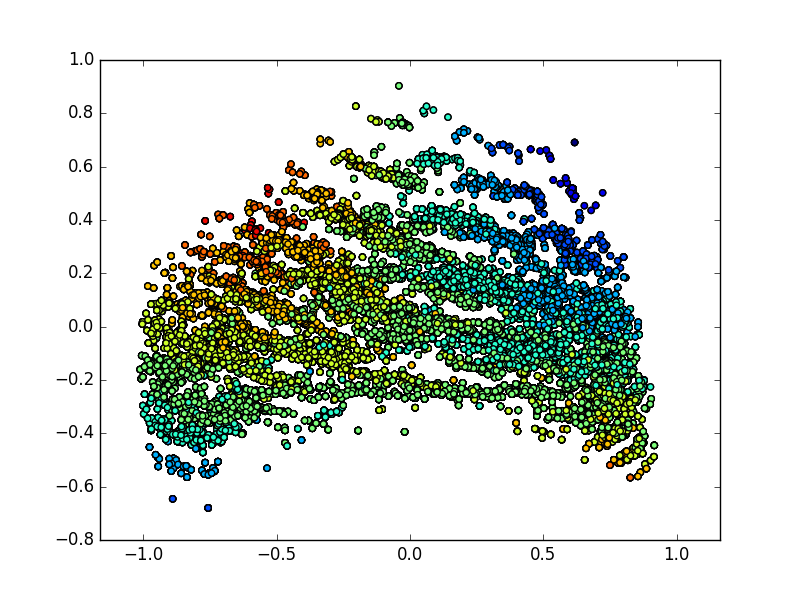}
\\ (a) Lateral Displacement
\label{fig:database_screenshot_nothing}
\end{minipage}
\hspace{\stretch{1}}
\begin{minipage}[t]{0.24\textwidth}
\centering
\includegraphics[width=\textwidth]{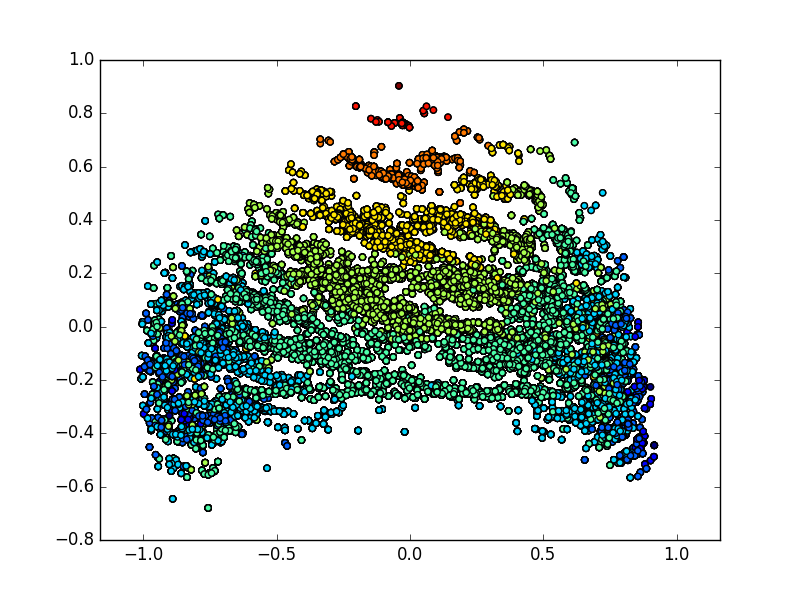}
\\ (b)  Longitudinal Displacement
\label{fig:database_screenshot_line}
\end{minipage}
\hspace{\stretch{1}}
\begin{minipage}[t]{0.24\textwidth}
\centering
\includegraphics[width=\textwidth]{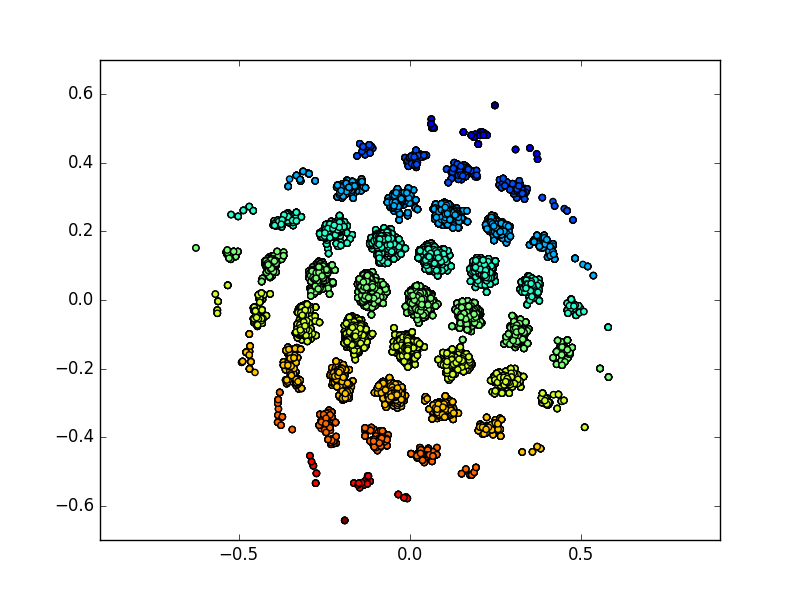}
\\ (c) Lateral Displacement (with inverse model)
\label{fig:database_screenshot_corner}
\end{minipage}
\hspace{\stretch{1}}
\begin{minipage}[t]{0.24\textwidth}
\centering
\includegraphics[width=\textwidth]{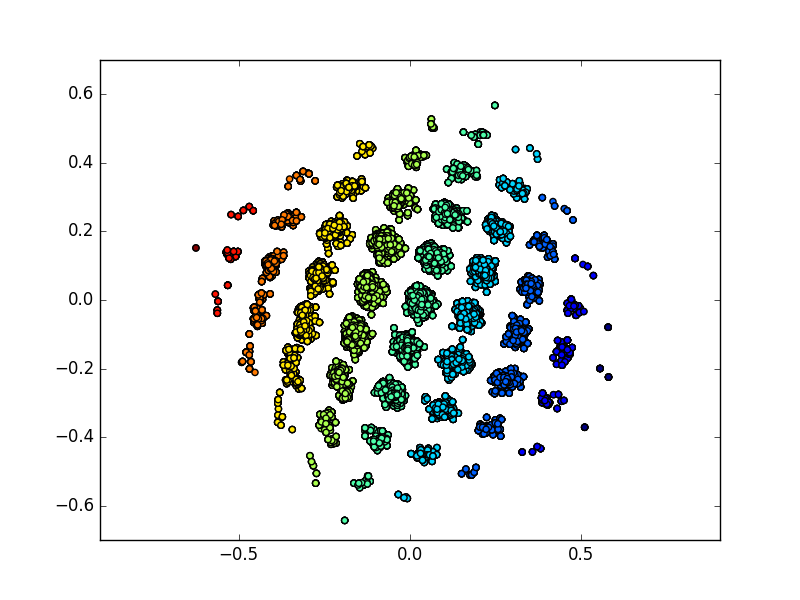}
\\ (d) Longitudinal Displacement (with inverse model)
\label{fig:database_screenshot_wallleft}
\end{minipage}

\caption{Projection of $\mathbf{h^m}$ usig PCA, color-coded for different values of the relative displacement of the agent in Cartesian coordinates}
\label{fig:baseline}
\end{figure*}

\begin{figure*}[h!]
\begin{minipage}[c]{0.15\textwidth}\centering
$\theta^{max}_{body} = \dfrac{2}{10}\pi$
\end{minipage}
\hspace{\stretch{1}}
\begin{minipage}[c]{0.25\textwidth}
\centering
\includegraphics[width=0.75\textwidth,trim={1.6cm 1.4cm 1.6cm 1.4cm},clip]{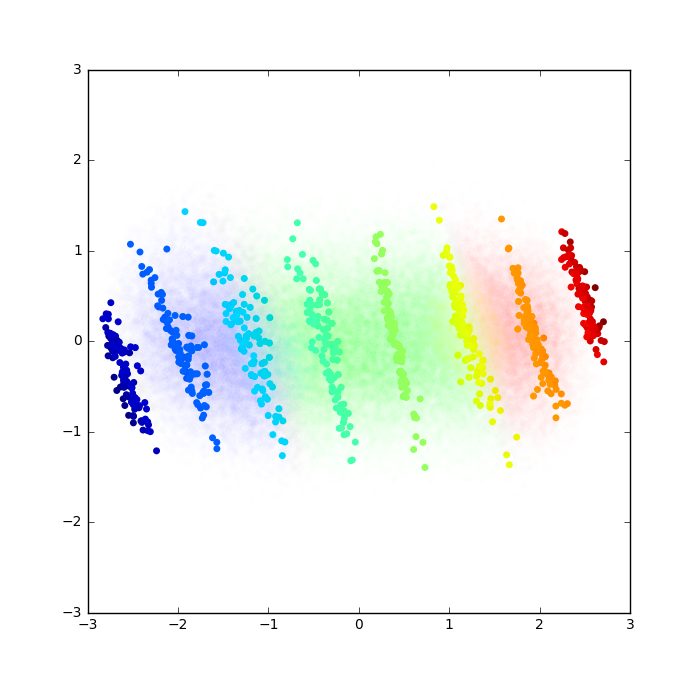}
\label{fig:forward_x_0_2}
\end{minipage}
\hspace{\stretch{1}}
\begin{minipage}[c]{0.25\textwidth}
\centering
\includegraphics[width=0.75\textwidth,trim={1.6cm 1.4cm 1.6cm 1.4cm},clip]{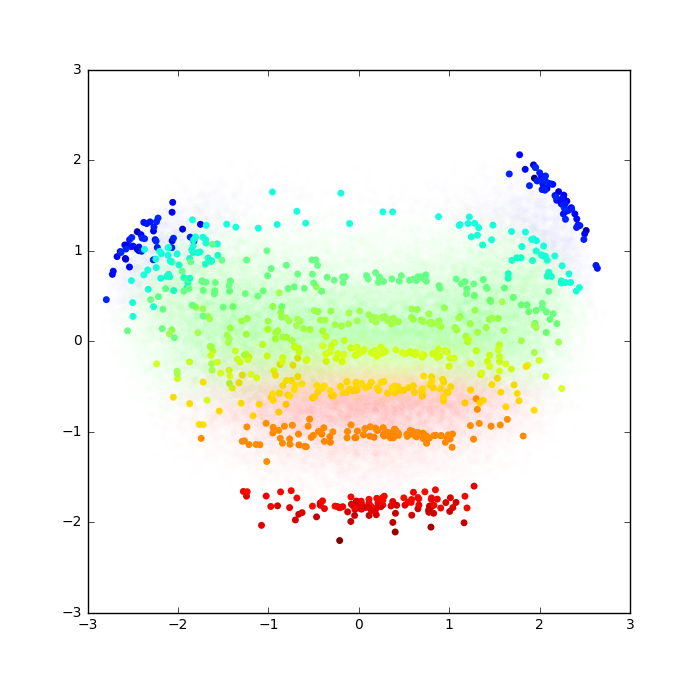}
\label{fig:forward_y_0_2}
\end{minipage}
\hspace{\stretch{1}}
\begin{minipage}[c]{0.25\textwidth}
\centering
\includegraphics[width=0.75\textwidth,trim={1.6cm 1.4cm 1.6cm 1.4cm},clip]{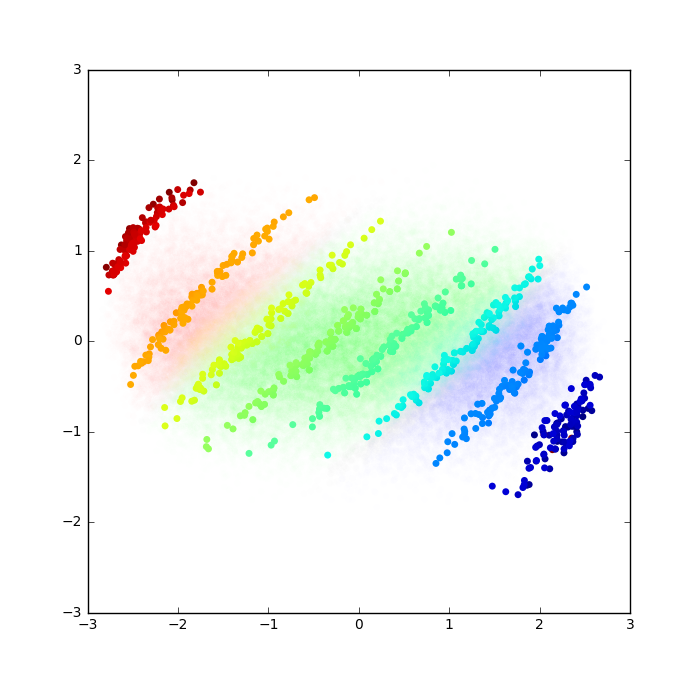}
\label{fig:forward_o_0_2}
\end{minipage}
\hspace{\stretch{1}}
\begin{minipage}[c]{0.15\textwidth}\centering
$\theta^{max}_{body} =\dfrac{4}{10}\pi$
\end{minipage}
\hspace{\stretch{1}}
\begin{minipage}[c]{0.25\textwidth}
\centering
\includegraphics[width=0.75\textwidth,trim={1.6cm 1.4cm 1.6cm 1.4cm},clip]{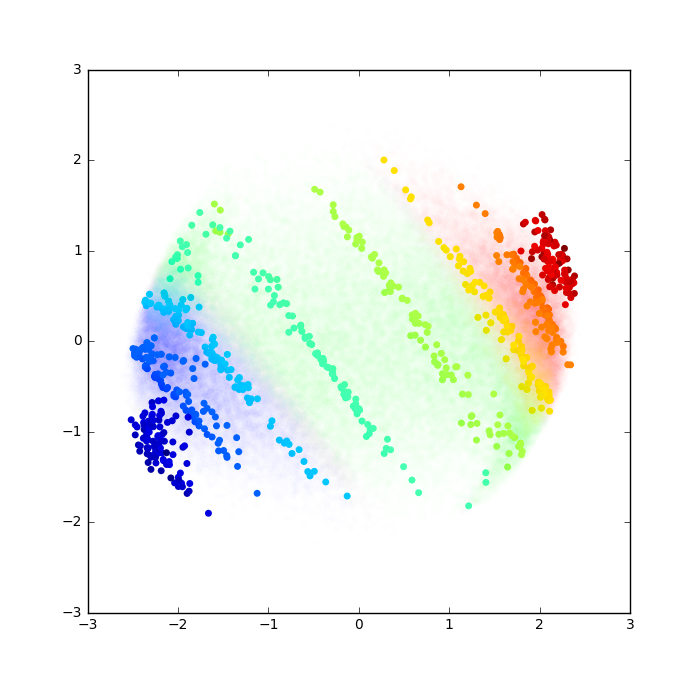}
\label{fig:forward_x_0_4}
\end{minipage}
\hspace{\stretch{1}}
\begin{minipage}[c]{0.25\textwidth}
\centering
\includegraphics[width=0.75\textwidth,trim={1.6cm 1.4cm 1.6cm 1.4cm},clip]{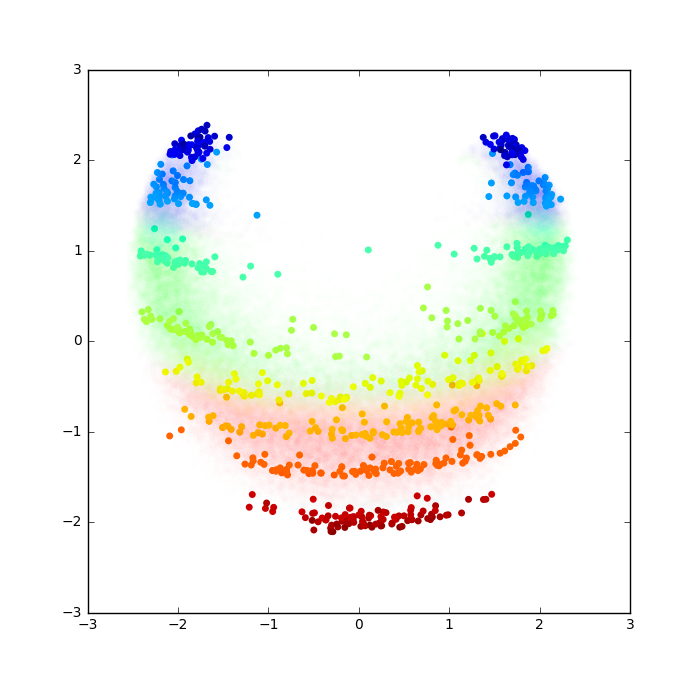}
\label{fig:forward_y_0_4}
\end{minipage}
\hspace{\stretch{1}}
\begin{minipage}[c]{0.25\textwidth}
\centering
\includegraphics[width=0.75\textwidth,trim={1.6cm 1.4cm 1.6cm 1.4cm},clip]{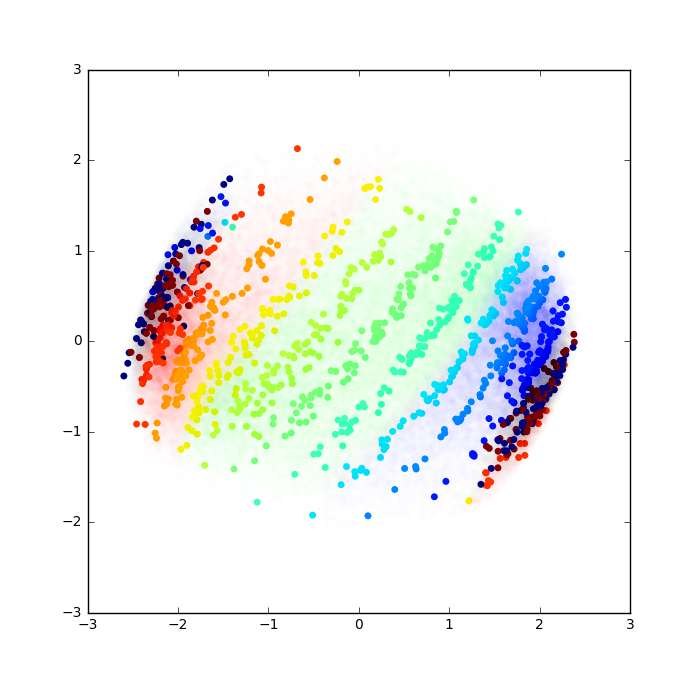}
\label{fig:forward_o_0_4}
\end{minipage}
\hspace{\stretch{1}}
\begin{minipage}[c]{0.15\textwidth}\centering
$\theta^{max}_{body} = \dfrac{8}{10}\pi$
\end{minipage}
\hspace{\stretch{1}}
\begin{minipage}[c]{0.25\textwidth}
\centering
\includegraphics[width=0.75\textwidth,trim={1.6cm 1.4cm 1.6cm 1.4cm},clip]{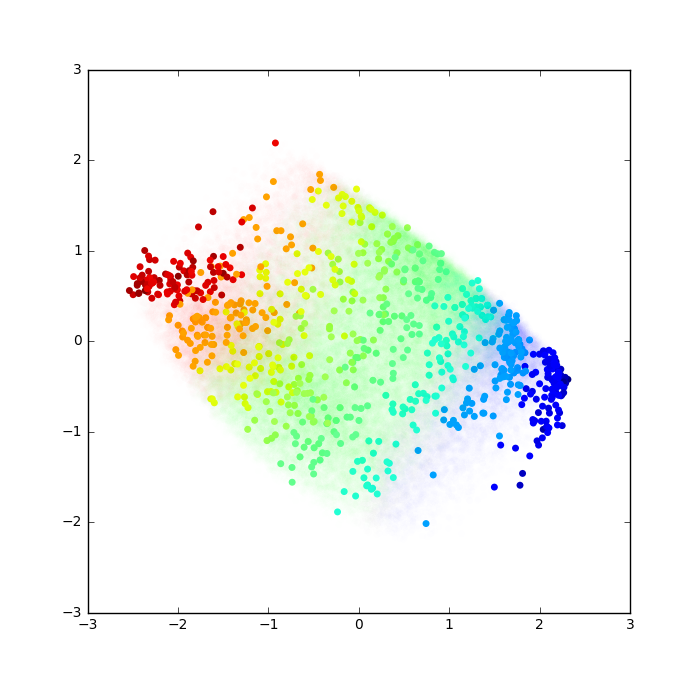}
\\  Longitudinal Displacement
\label{fig:forward_o_0_8}
\end{minipage}
\hspace{\stretch{1}}
\begin{minipage}[c]{0.25\textwidth}
\centering
\includegraphics[width=0.75\textwidth,trim={1.6cm 1.4cm 1.6cm 1.4cm},clip]{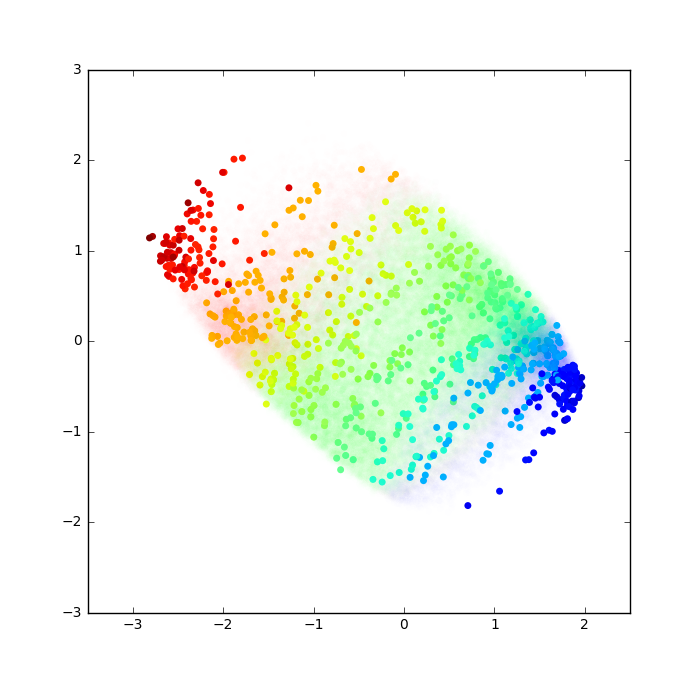}
\\  Lateral Displacement
\label{fig:forward_y_0_8}
\end{minipage}
\hspace{\stretch{1}}
\begin{minipage}[c]{0.25\textwidth}
\centering
\includegraphics[width=0.75\textwidth,trim={1.6cm 1.4cm 1.6cm 1.4cm},clip]{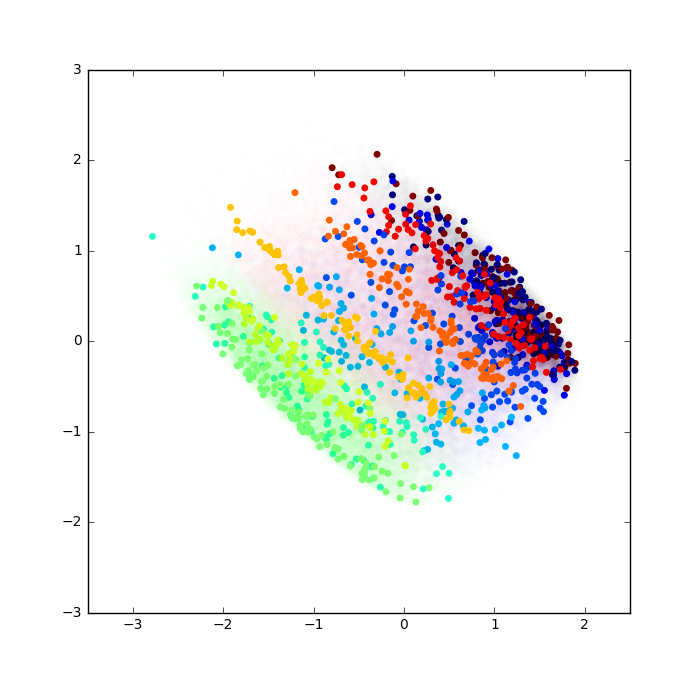}
\\ Change in Orientation
\label{fig:forward_o_0_8}
\end{minipage}
\caption{Projection of $\mathbf{h^m}$ using PCA, color-coded for different values of the relative displacement}
\label{fig:topo_forward}
\end{figure*}

\subsection{Baseline Environment}

We train our system on the Baseline Environment with 500 epochs of 50000 trajectories, using mini-batches of 1000 samples. We evaluated the prediction for sequences of 1 to 5 successive motor commands, and found that we reach perfect prediction for all motor sequences.  
By nature, the Baseline environment is completely deterministic and unambiguous, and it is therefore possible to predict with no error given that the prediction network has enough capacity to represent all possible sensorimotor transitions.

We can see in Fig.\ref{fig:baseline} (a) and (b) that the projection of $\mathbf{h^m}$ along the axis of the PCA is not organized topologically (we verified that the PCA is of dimension 2). We hypothesize that it is because the problem is simple enough for the network to represent sequences leading to similar displacement into different points in $\mathbf{h^m}$, while still being able to perform accurate prediction. We propose to add, for the case of the Baseline Environment only, an auxiliary task that forces the motor representation space to bring together different sequences leading to similar displacement. We add an additional prediction network which takes as input the present and future sensor values $ [ \mathbf{s}_t, \mathbf{s}_{t+T} ] $ and predicts $\mathbf{h^m}$ that led to this sensory change. This network, which learns an inverse model, is trained together with the proposed system by adding a prediction loss for  $\mathbf{h^m}$. It lead to the representation spaces presented in Fig.\ref{fig:baseline} (c) and (d).
On the Baseline example, learning the forward model lead to the organization of the representation space $\mathbf{h^m}$ into a topological space only if the inverse model is also learned. However, we will see in Sec.\ref{sec:exp_forward} and Sec.\ref{sec:exp_holo} that this additional network is not necessary when the sensorimotor space is continuous and rich.

\subsection{Forward Agent}\label{sec:exp_forward}

The Forward Agent performs sequences of $T=6$ motor commands in the continuous environment, as described in Sec.\ref{sec:forward_agent}, with $ \delta^{max}_{forward} = 1.0 $. We train different systems corresponding to different amplitudes of body rotation of the agent  $\theta^{max}_{body}$. 
First of all, we compare the results of the prediction in term of Mean Squared Error in Tab.\ref{table:prediction_theta}  for different values of $\theta^{max}_{body}$. We observe that the prediction scores are not affected by larger rotation angle.

In order to visualize the properties of the space $\mathbf{h^m}$ learned by our system, we project it along different axis of a PCA, and present the plots in Fig.\ref{fig:topo_forward}. In order to help visualization, most of the points are plotted with transparency. We plot points corresponding to particular ranges of displacement and rotation without transparency to help the visualization of the actual shape of the learned space.
As can be seen, the representation is organized along axis of variations which correspond to the actual dimensionality of displacement. Additionally, we observe that the subspace corresponding to the orientation of the agent curves when the angle of rotation becomes large. It describes a circular shape along which the different orientations are organized. 

\begin{figure*}[h!]
\begin{minipage}[c]{0.15\textwidth}\centering
$\theta^{max}_{body} = \dfrac{2}{10}\pi$
\end{minipage}
\hspace{\stretch{1}}
\begin{minipage}[c]{0.25\textwidth}
\centering
\includegraphics[width=0.75\textwidth,trim={1.6cm 1.4cm 1.6cm 1.4cm},clip]{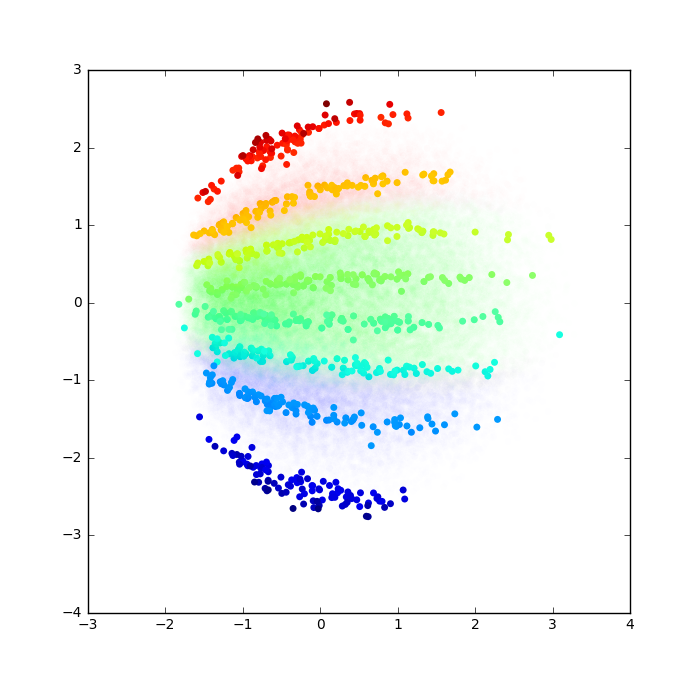}
\end{minipage}
\hspace{\stretch{1}}
\begin{minipage}[c]{0.25\textwidth}
\centering
\includegraphics[width=0.75\textwidth,trim={1.6cm 1.4cm 1.6cm 1.4cm},clip]{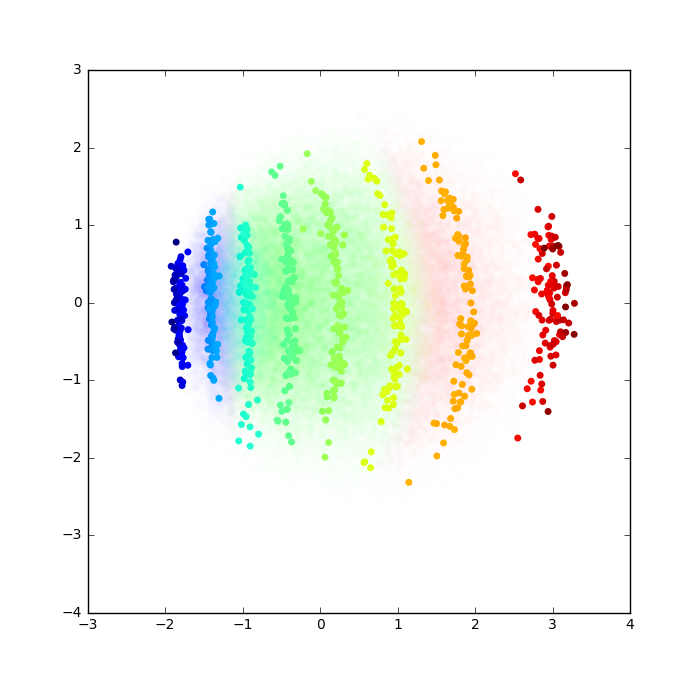}
\end{minipage}
\hspace{\stretch{1}}
\begin{minipage}[c]{0.25\textwidth}
\centering
\includegraphics[width=0.75\textwidth,trim={1.6cm 1.4cm 1.6cm 1.4cm},clip]{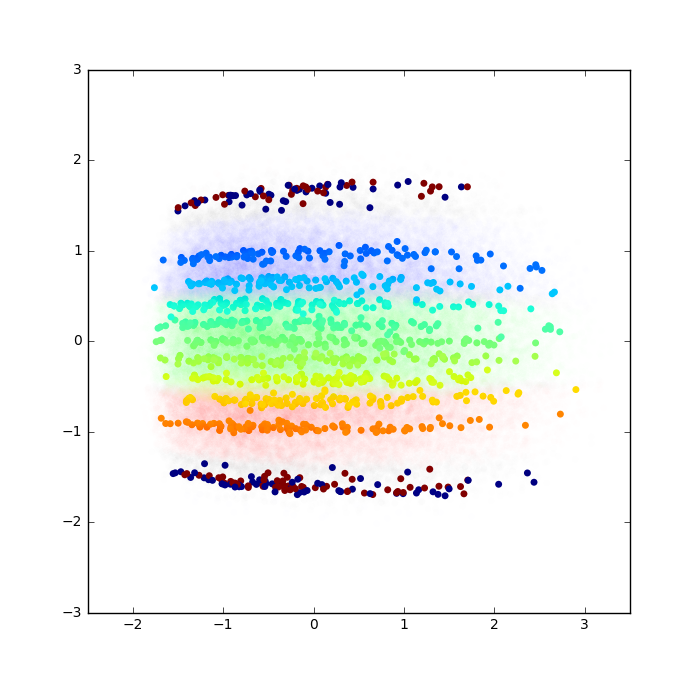}
\end{minipage}
\hspace{\stretch{1}}
\begin{minipage}[c]{0.15\textwidth}\centering
$\theta^{max}_{body} = \dfrac{4}{10}\pi$
\end{minipage}
\hspace{\stretch{1}}
\begin{minipage}[c]{0.25\textwidth}
\centering
\includegraphics[width=0.75\textwidth,trim={1.6cm 1.4cm 1.6cm 1.4cm},clip]{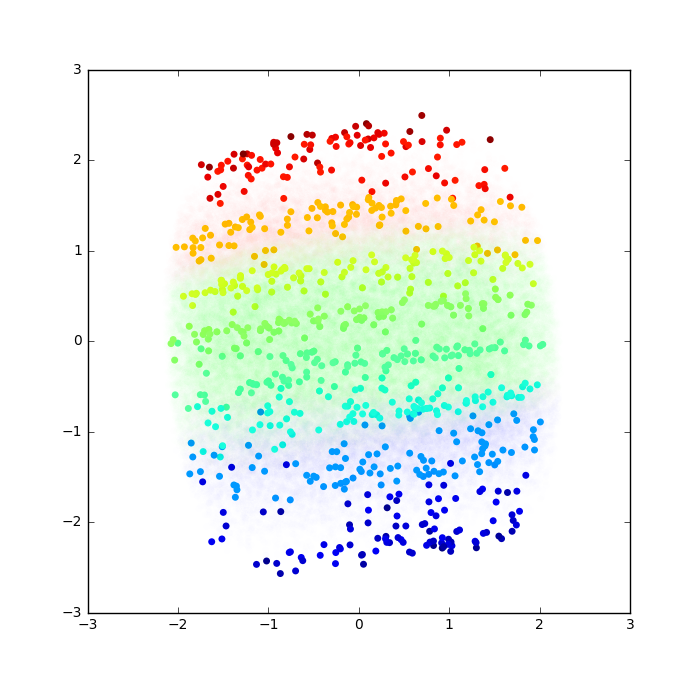}
\end{minipage}
\hspace{\stretch{1}}
\begin{minipage}[c]{0.25\textwidth}
\centering
\includegraphics[width=0.75\textwidth,trim={1.6cm 1.4cm 1.6cm 1.4cm},clip]{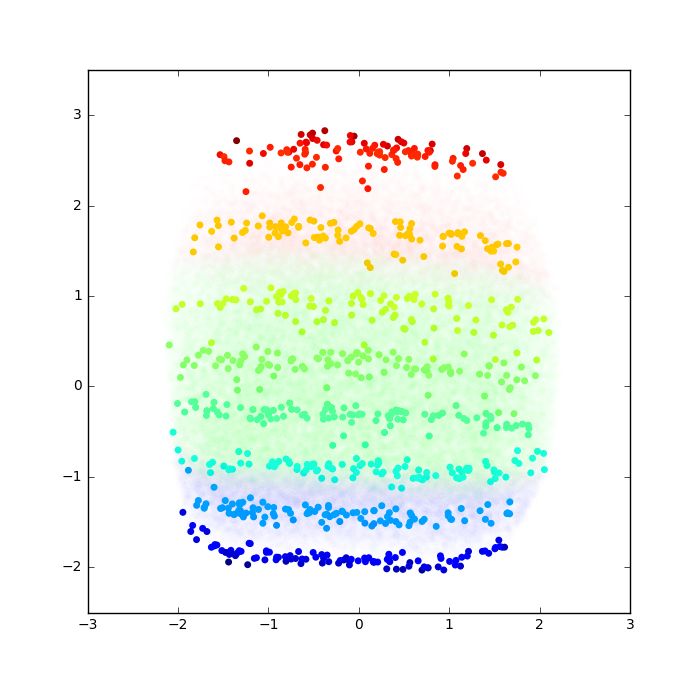}
\end{minipage}
\hspace{\stretch{1}}
\begin{minipage}[c]{0.25\textwidth}
\centering
\includegraphics[width=0.75\textwidth,trim={1.6cm 1.4cm 1.6cm 1.4cm},clip]{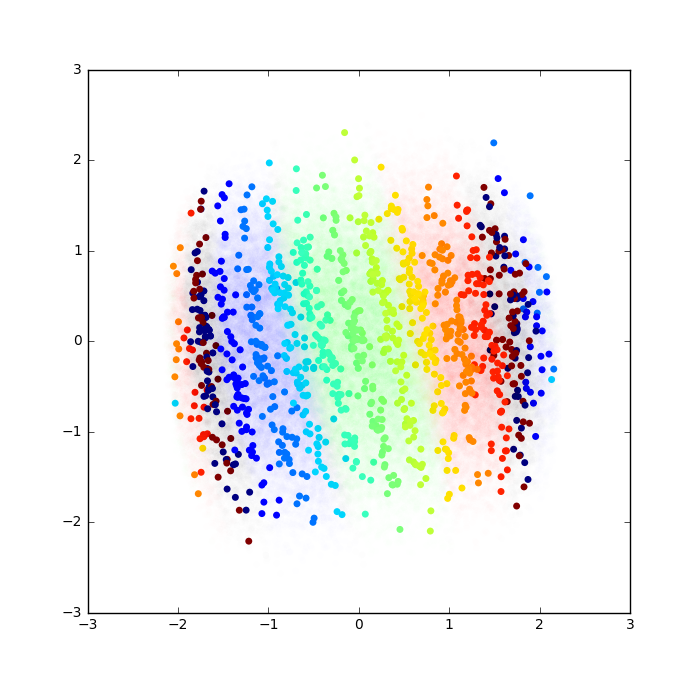}
\end{minipage}
\hspace{\stretch{1}}
\begin{minipage}[c]{0.15\textwidth}\centering
$\theta^{max}_{body} = \dfrac{8}{10}\pi$
\end{minipage}
\hspace{\stretch{1}}
\begin{minipage}[c]{0.25\textwidth}
\centering
\includegraphics[width=0.75\textwidth,trim={1.6cm 1.4cm 1.6cm 1.4cm},clip]{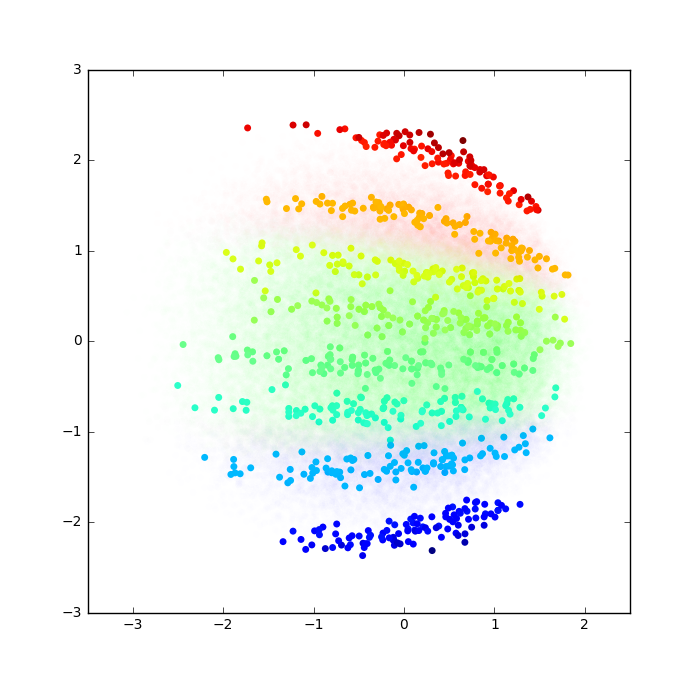}
\\ Lateral Displacement
\end{minipage}
\hspace{\stretch{1}}
\begin{minipage}[c]{0.25\textwidth}
\centering
\includegraphics[width=0.75\textwidth,trim={1.6cm 1.4cm 1.6cm 1.4cm},clip]{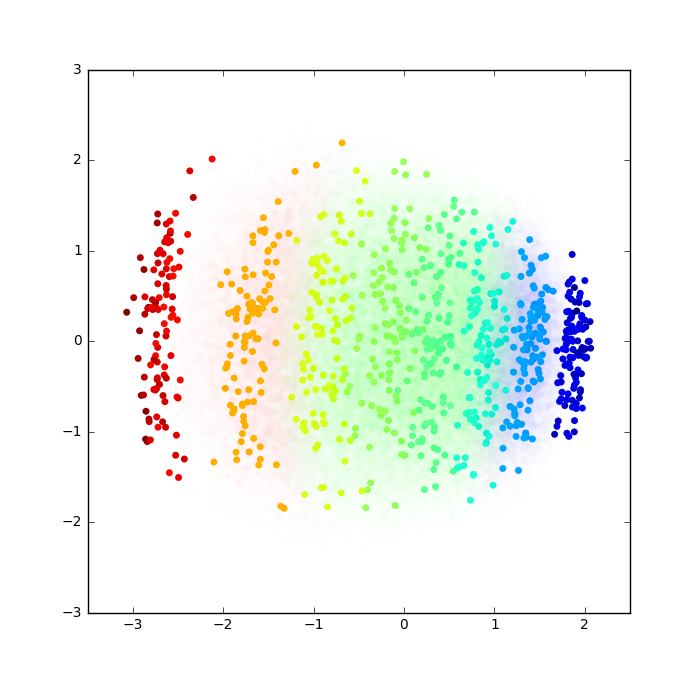}
\\ Longitudinal Displacement
\end{minipage}
\hspace{\stretch{1}}
\begin{minipage}[c]{0.25\textwidth}
\centering
\includegraphics[width=0.75\textwidth,trim={1.6cm 1.4cm 1.6cm 1.4cm},clip]{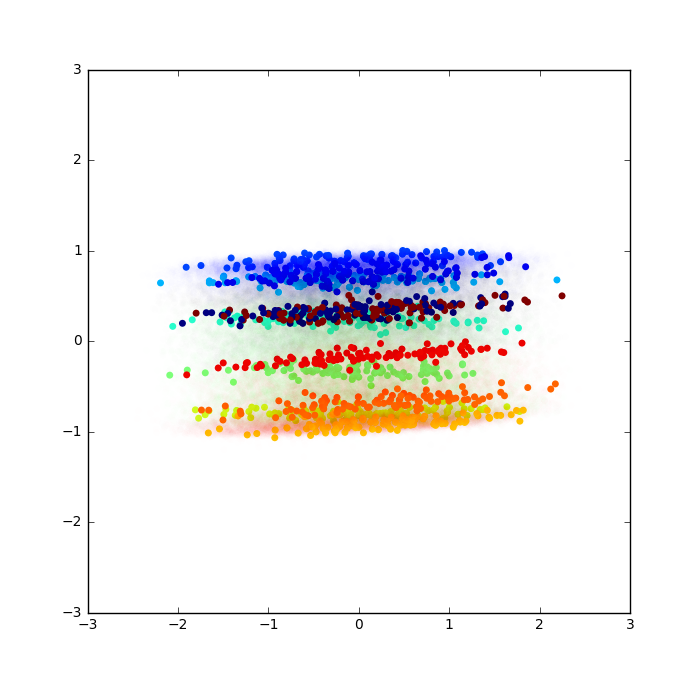}
\\ Orientation
\label{fig:holo}
\end{minipage}

\caption{Projection of $\mathbf{h^m}$ using PCA, color-coded for different values of the relative displacement of the agent in Cartesian coordinates X and Y}
\label{fig:forward}
\end{figure*}

\subsection{Holonomic Agent}\label{sec:exp_holo}

We perform similar experiments for the Holonomic Agent as for the Forward Agent. We sample sequences of $T=6$ movements in the continuous environment, as described in Sec.\ref{sec:holo_agent} with $ \delta^{max}_{long} = 1.0 $, $ \delta^{max}_{lat} = 1.0 $, and $ \theta^{max}_{head} = \dfrac{5}{10}\pi $ . In Tab.\ref{table:prediction_theta}, we present the results of prediction in terms of Mean Square Error for different values of $\theta^{max}_{body}$. We train each system with 5000 epochs of 2000 trajectories, using mini-batches of 100 samples.
Similarly to the experiments with the Forward Agent, the prediction seems to be unaffected by changes in $\theta^{max}_{body}$. 
In Fig.\ref{fig:holo}, we show visualizations of the space $\mathbf{h^m}$ projected on axis of its  PCA. Similarly to the Forward Agent, we observe that the representation space is flat for small values of $\theta^{max}_{body}$, but curves for larger values. 

\begin{table}[h!]
\centering
\begingroup
\setlength{\tabcolsep}{8pt} 
\renewcommand{\arraystretch}{2.5} 
\begin{tabular}{ | c | c  c  c  c  c c |}
\hline
  $\theta^{max}_{body}$ & $\dfrac{2}{10}\pi$ & $\dfrac{4}{10}\pi$ & $\dfrac{6}{10}\pi$ & $\dfrac{8}{10}\pi$ & $\pi$ & $\dfrac{12}{10}\pi$ \\
\hline
  Forward & 2.32 & 2.41 & 2.34 & 2.33 & 2.35 & 2.32\\
  Holonomic & 1.54 & 2.21 & 2.30 & 2.32 & 2.63 & 2.67\\  
\hline
\end{tabular}
\vspace{10pt}
\caption{MSE computed with 10e5 trajectories for Forward and Holonomic agent, for different values of  $\theta^{max}_{body}$ }
\label{table:prediction_theta}
\endgroup
\end{table}

\subsection{Explanation for the curvature of the representation space}

\begin{figure}[h!]
\begin{minipage}[c]{0.24\textwidth}
\centering
\includegraphics[width=0.99\textwidth,trim={1.6cm 1.4cm 1.6cm 1.4cm},clip]{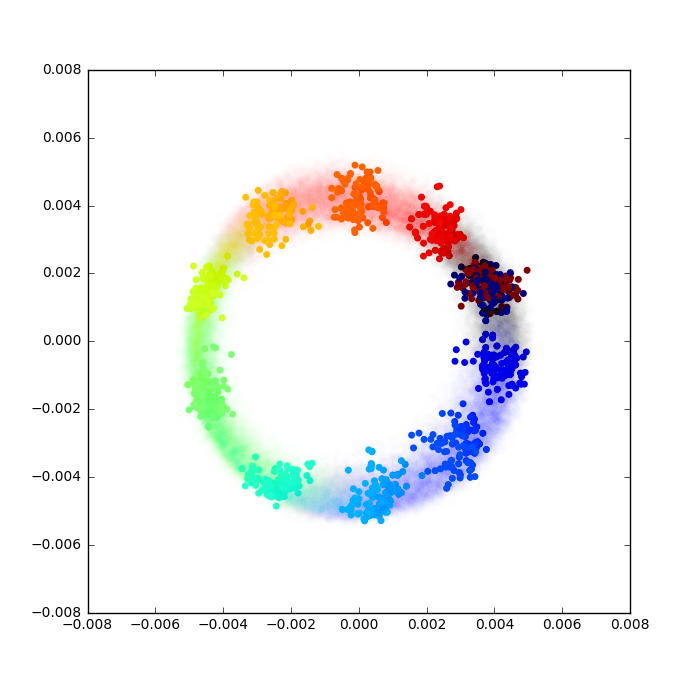}
\label{fig:forward_x_0_2}
\\Forward Agent
\end{minipage}
\begin{minipage}[c]{0.24\textwidth}
\centering
\includegraphics[width=0.99\textwidth,trim={1.6cm 1.4cm 1.6cm 1.4cm},clip]{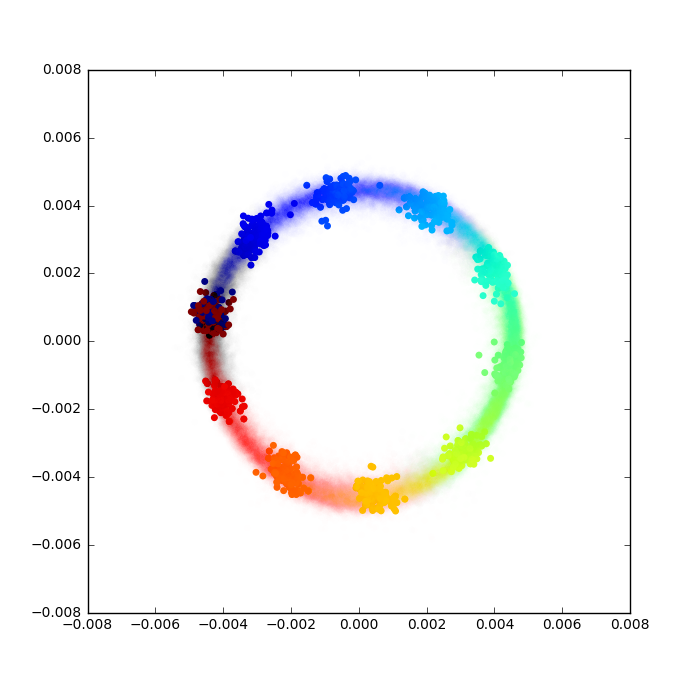}
\label{fig:forward_y_0_2}
\\Holonomic Agent
\end{minipage}
\caption{Projection of $\mathbf{h^m}$ using ICA, color-coded for different values of the relative displacement and change of orientation, for $\theta_{body}^{max} = \dfrac{8}{10}\pi $ }
\label{fig:ica}
\end{figure}

We notice on Fig.\ref{fig:forward} and Fig.\ref{fig:holo} that the space of motor sequences representation has a circular shape with regard to its orientation, for large values of $\theta^{max}_{body}$. An explanation might be that the agent experiences larger amplitude of body rotations, and observes more example of motor sequences which are equivalent in terms of displacement, but correspond to movements of rotation in different directions.
We use Independent Component Analysis (ICA) to extract independent directions of variation and display the orientation on axis of the ICA for the Forward and Holonomic Agent, for $\theta^{max}_{body} = 0.8$.
We present the result of the projection in Fig.\ref{fig:ica}, and show that our approach successfully collapses states which correspond to similar change in orientation.

\section{Conclusion}

We proposed a method for encoding sequences of motor commands into a compact space  which encodes for spatial displacement and change of orientation on its main axis of variation. This builds a topology of spatial displacement, for an agent moving in a simulated continuous environment. 
We showed that sensorimotor prediction can be used to guide the learning of this representation space, and illustrated on several examples that the shape of this representation depended on the type of motor commands and on the relative amplitude of the different modalities of displacement.
We hypothesize that the sensorimotor prediction learning exploits the continuity in sensory and motor space, and maintains this continuity in the representation space of motor sequences.

This approach to learn representations for spatial displacement can be useful in the framework of Developmental Robotics and open-ended learning. It is in particular relevant for the case when an agent needs to organize its sensorimotor experiences in space and time. Having access to this compact representation allows to use it as a basis to relate sensory experiences corresponding to different trajectories, but similar displacement, in an unambiguous fashion. Therefore, this representation can be used as a proxy for odometry and can be used to build a map of the environment without having access to actual odometry or absolute coordinates.
State Representation Learning is a very active field of research, mainly focused on encoding sensory states and use these states to solve reinforcement learning tasks. We think that navigation tasks in Reinforcement Learning could also benefit from learning representations of motor sequences.


\section{Future Works}

The study presented here is based on a simulation of an agent equipped with distance sensors. Distance sensors implicitly contain spatial information, which the sensorimotor learning can extract to build a representation of displacements. However, agents can not always rely on distance information. We want to study whether sensorimotor prediction can allow to extract information when the sensors do not explicitly contain any spatial information. For this purpose, we will apply our approach on vision sensors, where the agent would have to infer displacement from changes in visual representations.

Additionally, we will investigate the possibility to use the learned representation of displacement $\mathbf{h^m}$  as a latent space that the agent can control in order to generate sequences of motor commands. Learning this inverse model would allow to build a controller for the agent, and to generate motor commands that transform the current sensory values into desirable future sensory values.

Finally,  we want to apply the learned representations in the context of Reinforcement Learning, to solve a Navigation task. This would allow us to evaluate quantitatively the benefits of such representations in terms of generalization and transfer to new environments.

\newpage
\bibliographystyle{IEEEtran}
\bibliography{biblio}

\end{document}